\documentclass[10pt,letterpaper]{article}

\usepackage{cogsci}
\cogscifinalcopy 

\usepackage{pslatex}
\usepackage{apacite}
\usepackage{graphicx}
\usepackage{float}
\usepackage{linguex}
\usepackage{amsmath}
\usepackage{url}
\usepackage{xspace}

\usepackage{todonotes}

\newcommand{\dll}{\Delta\mbox{LogLik}}
\newcommand{\predictivePower}{psychometric predictive power\xspace}

\title{On the Predictive Power of Neural Language Models for Human Real-Time Comprehension Behavior}
 
\author{Ethan G. Wilcox$^1$, Jon Gauthier$^2$, Jennifer Hu$^2$, Peng Qian$^2$, \and Roger P.\ Levy$^2$ \\
  $^1$Department of Linguistics, Harvard University\\
  $^2$Department of Brain and Cognitive Sciences, Massachusetts Institute of Technology \\
  \texttt{\{jennhu,pqian,rplevy\}@mit.edu}\\
  \texttt{jon@gauthiers.net}, \texttt{wilcoxeg@g.harvard.edu}} 

\begin{document}

\maketitle

\begin{abstract}

Human reading behavior is tuned to the statistics of natural language: the time it takes human subjects to read a word can be predicted from estimates of the word's probability in context. 
However, it remains an open question what computational architecture best characterizes the expectations deployed in real time by humans that determine the behavioral signatures of reading.  Here we test over two dozen models, independently manipulating computational architecture and training dataset size, on how well their next-word expectations predict human reading time behavior on naturalistic text corpora.  Consistent with previous work, we find that across model architectures and training dataset sizes the relationship between word log-probability and reading time is (near-)linear.  We next evaluate how features of these models determine their psychometric predictive power, or ability to predict human reading behavior. In general, the better a model's next-word expectations (as measured by the traditional language modeling perplexity objective), the better its psychometric predictive power. However, we find nontrivial differences in psychometric predictive power across model architectures.  For any given perplexity, deep Transformer models and $n$-gram models generally show superior psychometric predictive power over LSTM or structurally supervised neural models, especially for eye movement data.  Finally, we compare models' psychometric predictive power to the depth of their syntactic knowledge, as measured by a battery of syntactic generalization tests developed using methods from controlled psycholinguistic experiments. Once perplexity is controlled for, we find no significant relationship between syntactic knowledge and predictive power. These results suggest that, at least for the present state of natural language technology, different approaches may be required to best model human real-time language comprehension behavior in naturalistic reading versus behavior for controlled linguistic materials designed for targeted probing of syntactic knowledge.

\end{abstract}

\section{Introduction}

\renewcommand{\thefootnote}{}
\footnotetext{Scripts and data can be found online at \url{https://github.com/wilcoxeg/neural-networks-read-times}.}
\renewcommand{\thefootnote}{\arabic{footnote}}
A large body of evidence suggests that humans are expectation-based language processors, insofar as real-time language comprehension involves making predictions about upcoming material \cite{levy2008expectation, hale2001probabilistic}. One strong piece of evidence supporting this view comes from the domain of computational modeling, where next-word log probabilities from statistical language models (LMs) turn out to correlate well with online processing measures---that is, to have good \emph{\predictivePower}---including gaze duration in eye-tracking studies and self-paced reading times \cite{smith2013effect}, and the N400 measure in EEG studies \cite{frank-etal:2015erp}. Crucially, as statistical LMs improve on the broad-coverage objective function of perplexity (i.e.\ as they get better at predicting the next word given its context), so too do they improve at predicting real-time processing data \cite{goodkind2018predictive}.

Many of the previous studies linking information-theoretic measures and human psychometric data  were conducted using $n$-gram models, which track local word co-occurrences and are blind to information outside of the $n$-gram window. Recently, however, neural network models such as Long Short-Term Memory Recurrent Neural Networks (LSTM-RNNs; \citeNP{Elman:1990, Hochreiter:Schmidhuber:1997}) and Transformers \cite{Vaswani:et-al:2017} have set new standards in natural language processing, achieving state-of-the-art perplexity results. We present a broad evaluation of these modern neural network models as predictors of human reading behavior, testing the influence of both model inductive bias and the scale of training data provided to the model.

One important unanswered question involves the role of syntactic knowledge in the link between statistical models and real-time processing. Experimental evidence, such as studies of garden-path effects, demonstrates that humans deploy hierarchically structured representations to drive predictions about upcoming material \cite{stowe:1986parsing,staub-clifton:2006}. This suggests that language models with similar syntactic capacity --- represented implicitly or explicitly --- may be the best candidates for predicting human processing data.
However, results from computational modeling paint a complicated story: while \citeA{frank2011insensitivity} found that models without explicit hierarchical structure are best at predicting human reading times of naturalistic text, a follow-up study conducted by \citeA{fossum2012sequential} argued that perplexity, not inductive bias or syntactic capacity, was the primary factor in determining a the ability of NLP models of that time to predict human reading times.  The more recent work of \citeA{goodkind2018predictive}, \citeA{aurnhammer2019comparing}, and \citeA{merkx2020comparing} confirm the general finding that perplexity is the primary determinant of model fit to human comprehension measures, but also find differences among model architectures once perplexity is controlled for.

Here we contribute to this emerging picture through a scaled-up and carefully controlled assessment of language models' ability to predict measures of human reading behavior. Following \citeA{hu2020assessing}, we train a fleet of neural-network language models varying both in inductive bias (from sequential LSTMs to syntax-aware recurrent models) and in the amount of data provided to them at training time.
We evaluate models' \predictivePower for human reading times on three online processing datasets: the Dundee eye-tracking corpus \cite{kennedy2003dundee}, selections from the Brown corpus and the Natural Stories self-paced reading time corpus \cite{futrell2017natural}. Across model architectures and training datasets, our results broadly confirm the strong linear relationship between surprisal (or negative log probability) and reading time originally documented by \citeA{smith-levy:2008,smith2013effect} and confirmed by \citeA{goodkind2018predictive}.  Like previous studies, we also find a generally positive relationship between a model's next-word prediction accuracy and its ability to predict human reading times, supporting the findings of \citeA{goodkind2018predictive} on a broad set of neural network models. Beyond the role of perplexity, we find that deep Transformer models demonstrate the best \predictivePower, and $n$-gram models achieve greater \predictivePower than would be expected based on their perplexity.

We next address the issue of syntactic knowledge.
Rather than positing a binary distinction between ``hierarchical'' and ``non-hierarchical'' models, we draw on recent work in language model evaluation to quantify models' syntactic knowledge at a finer grain \cite{hu2020assessing}. We compare each models' \predictivePower against this measure of syntactic knowledge. After controlling for a model's next-word prediction accuracy, we find that syntactic knowledge does not explain significant variance in a model's \predictivePower.


\section{Methods}

\subsection{Models}

We train a fleet of 
language models, each providing an estimate of word probability in context. The function of each language model is to predict the next token in a corpus $x_i$ conditioned on its preceding context $x_{j < i}$, producing a probability distribution $P_\text{model}(x_i \mid x_{j < i})$. Our fleet contains four major architectural variants:

\begin{itemize}
    \item \textbf{LSTM-RNNs} are recurrent neural networks with Long Short-Term Memory units \cite{Hochreiter:Schmidhuber:1997}. We employ the boilerplate PyTorch implementation \cite{paszke2017automatic}.
    \item \textbf{Recurrent Neural Network Grammars} \cite<RNNGs;>{Dyer:et-al:2016} model the joint probability of a sequence of words as well as its syntactic structure. RNNGs are supervised during training with Penn Treebank-style constituency parses \cite{Marcus:et-al:1993}.
    \item \textbf{Transformers} are deep neural networks which stack layers of self-attention mechanisms above word embedding representations, which have recently achieved state-of-the-art performance on language modeling and set a new standard for pretrained sentence encoding in natural language processing. We train the GPT-2 Transformer architecture \cite{radford2019language} from scratch on our own corpora.
    \item \textbf{$n$-gram:} We train a $5$-gram model with Kneser-Ney smoothing, using the SRILM language modeling toolkit \cite{stolcke2002srilm}.
\end{itemize}

Following \citeA{hu2020assessing}, we train each model on four corpora of varying sizes drawn from the Brown Laboratory for Linguistic Information Processing (BLLIP) corpus \cite{charniak2000bllip}. The corpora are sampled such that the training set of each corpus is a subset of each larger corpus. The four corpora are BLLIP-XS (40K sentences, 100K tokens); BLLIP-SM (200K sentences, 5M tokens); BLLIP-MD (600K sentences, 14M tokens); and BLLIP-LG (2M sentences, 42M tokens). We trained 1--3 random seeds of each model architecture and training corpus. 

While the majority of the models tested here make predictions at the word level, 
some of our Transformers constitute a notable exception. These models instead make predictions at the sub-word level, using a byte-pair encoding \cite<BPE;>{sennrich2015neural}, which decomposes common word substrings into independent tokens. Models using this encoding can thus represent sublexical co-occurrence information. For the purposes of this paper, one of the most important possible effects of this sub-word representation may be that it supports well-tuned word probability estimates even for very rare or unknown words. We train Transformer models using both this BPE representation and standard word-level representations on the corpora mentioned above.


These language models are trained to minimize the perplexity of a corpus:
$\text{PPL}(\text{model}) = \left( \prod_i P_\text{model}(\text{word}_i \mid \text{words}_{j < i}) \right)^{-\frac 1 N}$. Lower perplexity values correspond to language models that make more accurate next-word predictions.%
\footnote{For models with sub-word representations, we define the probability of a word as the joint probability of its constituent subwords, following the chain rule.}
As perplexity is interpretable only in the context of a specific vocabulary (i.e., over a space of possible next words), perplexity measures are only comparable given a fixed reference vocabulary. However, if a model trained on a larger vocabulary has a better perplexity than a model trained on a smaller vocabulary, we can confidently say it is a better predictive model. This is generally the trend we find: models trained on larger corpora achieve better perplexity measures, despite being forced to predict over larger vocabularies. Nonetheless, most of our analyses in this paper will be comparing models with the same reference vocabulary to avoid this issue.

\subsection{Psychometric predictive power}

Following previous work \cite{frank2011insensitivity,fossum2012sequential,goodkind2018predictive}, we assess a model's \predictivePower (termed "psychological accuracy" by \citeauthor{frank2011insensitivity}) by asking how well its word-by-word surprisal estimates of the model can explain various psychometric measures of how subjects read individual words, after controlling for other features known to influence reading behavior, such as the length and frequency of words.

We draw psychometric data from three datasets across two measurement modalities of real-time human language comprehension: eye-tracking data from the Dundee corpus \cite{kennedy2003dundee}; self-paced reading data from selections from the Brown corpus of American English (as reported in \citeA{smith2013effect}); and self-paced reading data (herein SPRT) from the Natural Stories corpus \cite{futrell2017natural}. The Natural Stories corpus was explicitly designed to include syntactic constructions that are relatively rare in both spoken and written English, such as object-extracted relative clauses, topicalization, and long-distance dependencies.

For each language model, we fit regression models which predict these psychometric data averaged across experimental subjects. (For the Dundee eye-tracking corpus, we predict the average gaze duration by subject for each word.) Our regression models combine model-specific and model-invariant features of words. The main predictor of interest is word surprisal, or the negative logarithm of word probability: $S_\text{model}(x_i) = -\log_2 P_\text{model}(x_i \mid x_{j < i})$. For each word read by a human subject, we extract the context-specific surprisal of the word and the previous word (or the previous 3 words for SPRT) from a language model. The previous word estimates are included due to known spillover effects in both measurement paradigms \cite{smith2013effect}. We combine these surprisal estimates with model-invariant and context-invariant features of the current and previous word (or previous 3 words for SPRT) as control predictors: its length in characters, and its log-frequency (or log-unigram-probability).\footnote{Word frequencies were measured from the larger Wikitext-2 corpus \cite{Merity:et-al:2017}.}

We evaluate each regression model relative to a baseline model, which attempts to predict the same human psychometric data from just the control features. For each language model, we compute its \predictivePower by calculating the mean by-token difference in log-likelihood of the response variable between the two models, which we refer to as $\dll$. A positive $\dll$ value indicates that a language model's surprisal estimates lead to more accurate predictions of human reading behavior over the baseline model.

We repeat the above analyses with both generalized additive models (GAMs) and linear regression.\footnote{\label{ft:gam}The R command to run the eye-tracking model was: \texttt{read-time $\sim$ s(surp, bs = "cr", k = 20) + s(prev.surp, bs = "cr", k = 20) + te(freq, len, bs = "cr") + te(prev.freq, prev.len, bs = "cr")} for the GAM model and \texttt{psychometric $\sim$ surprisal + prev\_surp + prev2\_surp + prev3\_surp + freq * len + prev\_freq * prev\_len + prev2\_freq * prev2\_len + prev3\_len * prev3\_freq} for the linear model.} Qualitative results were similar with both approaches; unless otherwise noted we report the linear regression results in figures and statistical tests.

\begin{figure*}[ht]
    \centering
    \begin{minipage}{\textwidth}
    \includegraphics[width=\textwidth]{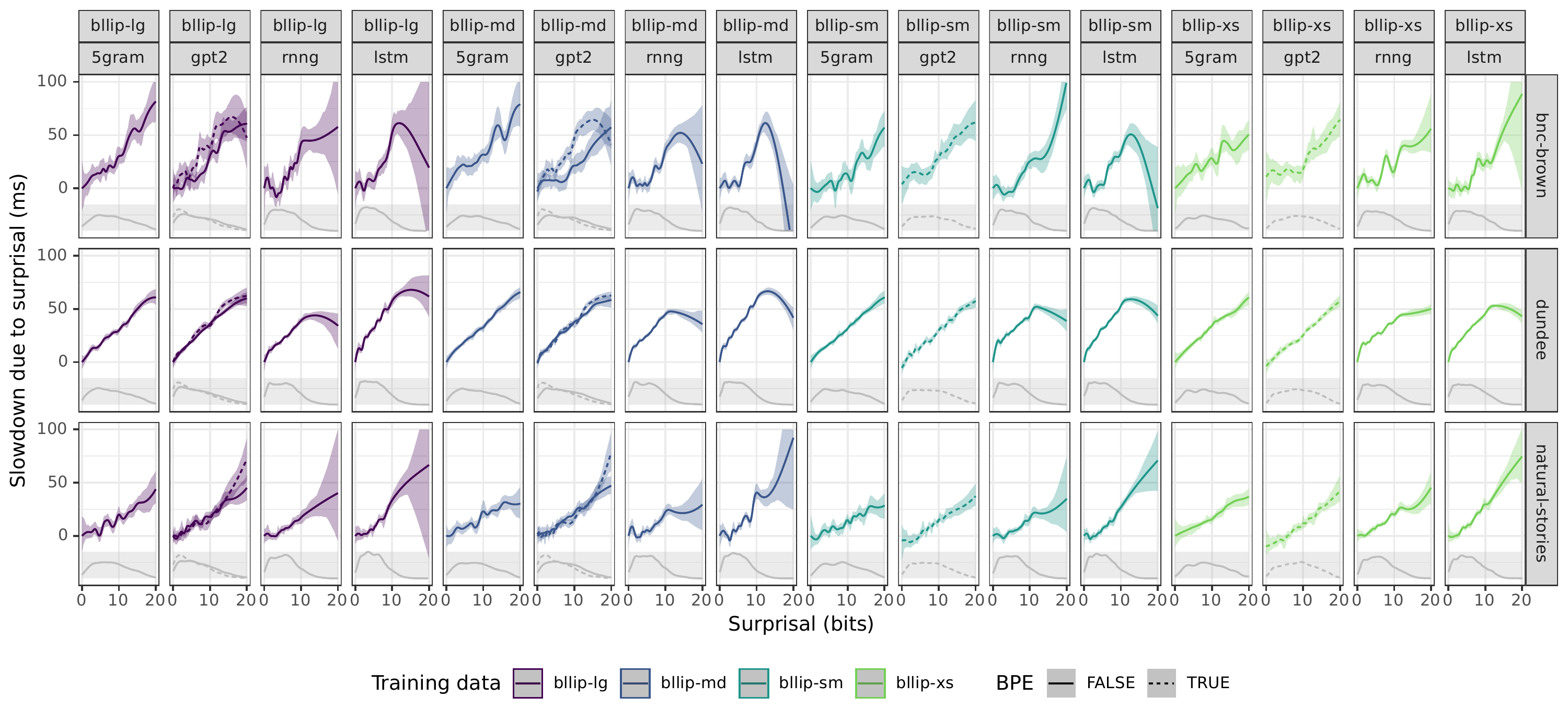}
    
    \end{minipage}
    \caption{The relationship between surprisal and reading time. Lines are regressions from fitted GAM models, using only context sensitive predictors (i.e. surprisal of the current and previous words) to derive estimates. Shaded regions are bootstrapped 95\% confidence intervals. Density plots of model-assigned surprisal values are below each fit. The GAM fits for the GPT-2 BPE models (\texttt{gptbpe}, rightmost column) pool surprisal estimates from models trained on all corpora.}
    \label{fig:surp_corr}
\end{figure*}


Our methods differ from \citeA{goodkind2018predictive} in two respects: First, instead of reporting the difference in joint log-likelihood of the entire dataset, we report the mean difference in log-likelihood between the baseline model and the predictive model on each individual token. Because the three corpora tested in this paper are very different in size and composition, the joint log-likelihood cannot be used to compare \predictivePower results across testing corpora.
The second key difference is that, whereas \citeA{goodkind2018predictive} report $\dll$ of the model on the training data, we report mean per-word $\dll$ of the model on held-out test data, averaged over 10-fold cross validation, allowing us to conduct analyses using GAM fits while guarding against overfitting.

\subsection{Syntactic Generalization score} \label{sec:syntactic-tests}

In order to assess the syntactic capabilities of each model, we report its score on the set of 34 targeted syntactic tests presented in \citeA{hu2020assessing}, which follow paradigms developed in \citeA{marvin-linzen:2018-targeted}, \citeA{futrell-etal:2018-arxiv-rnns-psycholinguistic-subjects}, and other recent papers on controlled psycholinguistics-style testing for grammatical knowledge. Each test is designed to probe whether the neural model has learned a particular aspect of English syntax by examining its behavior across minimally different sentence pairs. For example, \citeA{marvin-linzen:2018-targeted} assess whether a model has learned subject--verb number agreement by evaluating the model's behavior on a construction such as \textit{The keys to the cabinet are/is...}. If the model has learned the proper grammatical generalization regarding subject-verb number agreement, then it should assign lower probability to the ungrammatical continuation \textit{is} compared to the grammatical \textit{are}, conditioned on the fixed prefix \textit{The keys to the cabinet}.

Each individual syntactic test comprises between 20--30 test items, with each item used in multiple experimental conditions (generally 4, occasionally 2). In order for models to get the test item correct, their predictions must satisfy a set of inequality criteria among surprisals of regions of the sentence in each experimental condition. For example, following the logic described above, for subject--verb number agreement, a model must succeed at both of: (i) when the head noun of the subject NP is singular, the singular verb \textit{is} should be more likely than the plural verb \textit{are}; and (ii) when the head noun of the subject NP is plural, the plural verb \textit{are} should be more likely than the singular verb \textit{is}. 
This design ensures that models will be unable to get high scores by relying on simple heuristics, such as a broad preference for plural verbs.  We report models' mean by-test accuracy as its Syntactic Generalization (SG) score, which ranges from 0 to 1 with chance being $\sim$0.25.

\section{Results}

\subsection{Surprisal vs. Reading Times}

Figure \ref{fig:surp_corr} shows the relationship between language model surprisals and human reading times for all models and corpora.
Lines are fits from generalized additive models (trained using the formula described in Footnote \ref{ft:gam}), with only context sensitive predictors (i.e. surprisal of the current word and previous words) used to derive estimates. They show the contribution of surprisal on reading time separate from word length and word frequency. Although there is some variance based on model architecture and training corpus, overall we find a linear relationship holds for most of the models tested.



\subsection{Predictive Power vs. Perplexity}

\begin{figure*}[ht]
    \centering
    \includegraphics[width=0.99\textwidth]{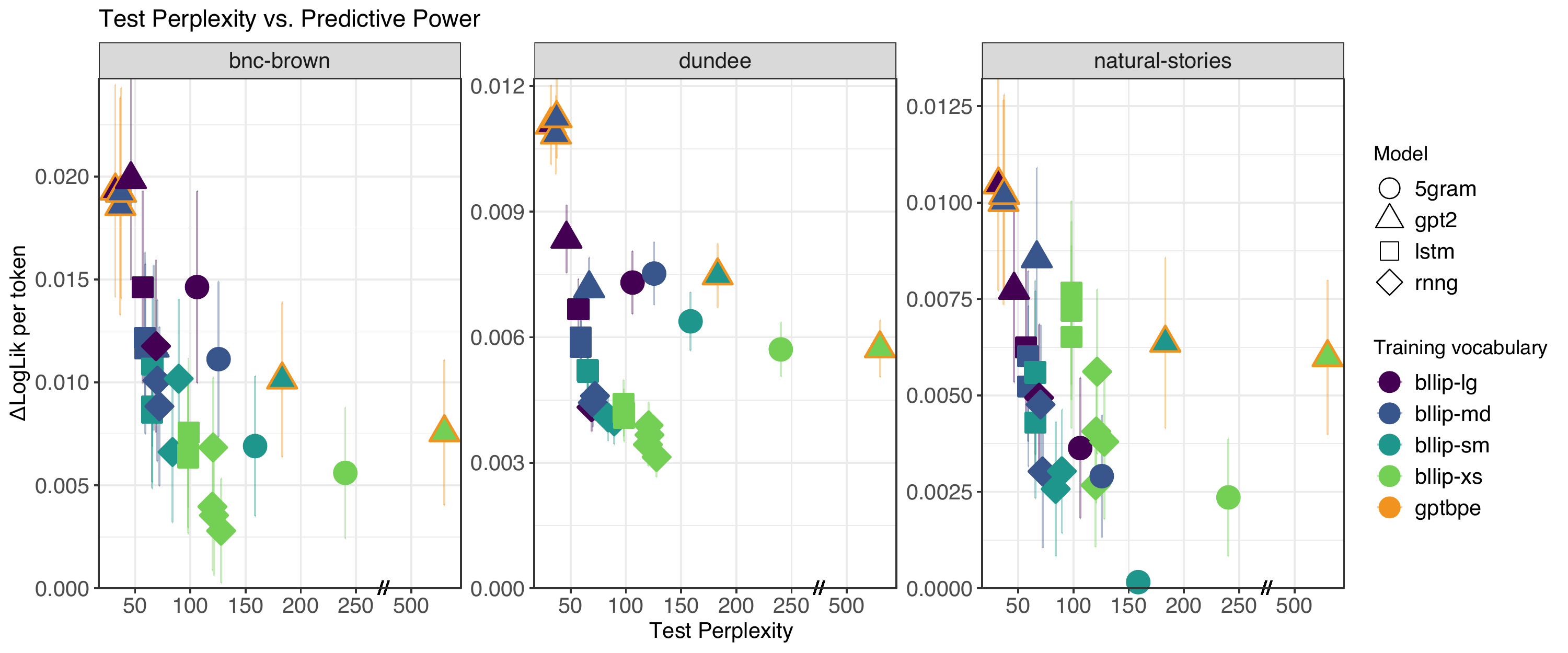}
    \caption{Relationship between predictive power ($\dll$) and model perplexity. Error bars are standard errors of by-fold mean $\dll$ per token, using 10-fold cross validation. As model perplexity decreases, predictive power increases for all test corpora.}
    \label{fig:ppl_loglik}
\end{figure*}

The relationship between \predictivePower and perplexity is shown in Figure \ref{fig:ppl_loglik}. 
Error bars denote the standard error of by-fold mean $\dll$ per token, estimated by 10-fold cross validation.
If better language models are better predictors of human processing time, we would expect a negative correlation between $\dll$ and perplexity, which is visually evident for all three testing corpora. On average, the Brown testing corpus shows slightly higher $\dll$, but also higher variance across the 10-fold split.

In order to assess the relationship between perplexity and \predictivePower, we fit a mixed-effects regression model to predict $\dll$ from language model perplexity within each training vocabulary, with random intercepts by test corpus and model architecture. We find a significant effect of perplexity on $\dll$ within each training vocabulary ($p<0.01$), except for in the BLLIP-LG training data, where $p=0.07$.
We take these results to indicate that the relationship found in \citeA{frank2011insensitivity,fossum2012sequential}, and \citeA{goodkind2018predictive} between a model's \predictivePower
and its test perplexity holds for a range of contemporary state-of-the-art models, and for perplexity scores in the 30-100 range. However, whereas \citeA{goodkind2018predictive} find a strongly linear relationship between perplexity and $\dll$, our results are a bit more complicated: While there is a monotonic relationship between $\dll$ and perplexity, this may look more or less linear depending on the model class. For example, focusing on the $n$-gram models tested on the Dundee corpus, the relationship appears strongly linear across the 100--250 perplexity range. However, focusing on the neural models in the 30--100 perplexity range, the relationship appears more exponential, with stronger $\dll$ gains between models in the lower perplexity range.

While all three testing corpora show a relationship between perplexity and $\dll$, we also find an effect of model class for Brown and Dundee. Here, the $n$-gram models demonstrate predictive power comparable to the neural models despite much poorer perplexity scores. This is especially evident for the BLLIP-SM and XS models tested on the Dundee corpus. While the $n$-gram models' perplexity is 2$\times$ that of the neural models, they achieve \textit{higher} average $\dll$. While surprising, this result accords with the findings presented in \citeA{goodkind2018predictive}, who find their LSTM model to underperform relative to their $n$-gram models.\footnote{The LSTM model in that paper's Figure 1 is the only model that falls outside the regression's 95\% confidence interval.}

\subsection{Psychometric Predictive Power vs. Syntactic Generalization}

\begin{figure*}[ht]
    \centering
    \begin{minipage}{0.23\textwidth}
    \includegraphics[width=\textwidth]{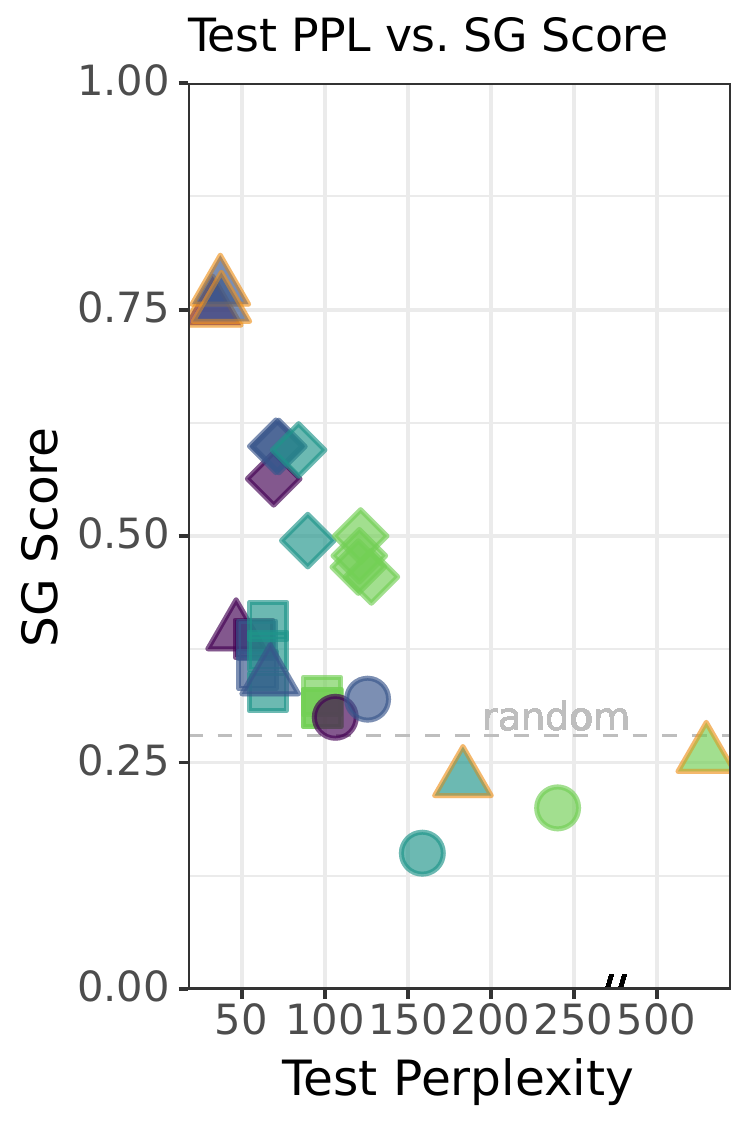}
    \vspace{-0.65cm}
    \caption{Range of SG scores achieved by models plotted against perplexity.}
    \label{fig:sg_ppl}
    \end{minipage}
    \hspace{0.1cm}
    \begin{minipage}{0.73\textwidth}
    \includegraphics[width=\textwidth]{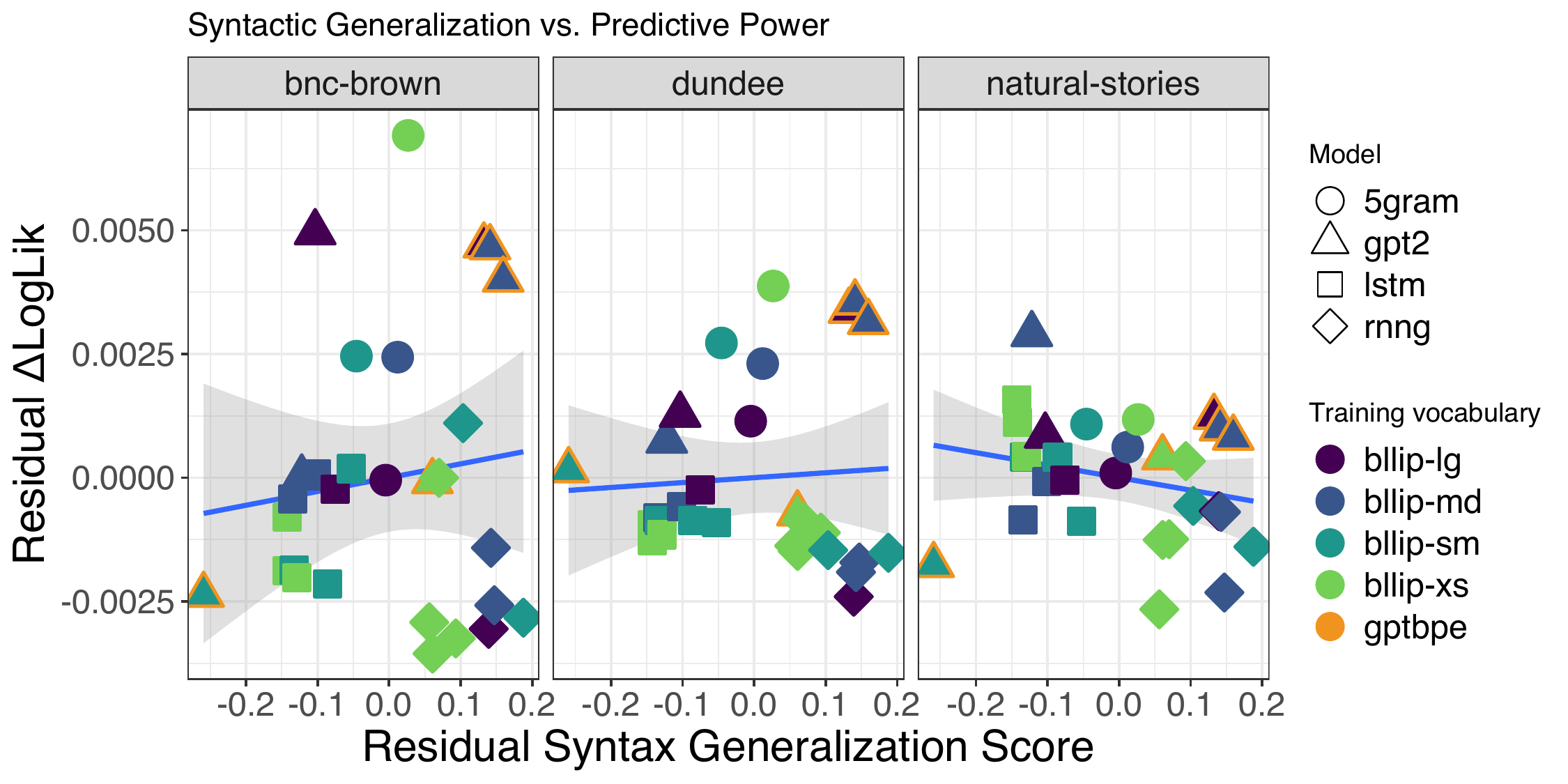}
    \vspace{-0.8cm}
    \caption{Lack of relationship between models' predictive power ($\dll$) and Syntactic Generalization score, both residualized with respect to model perplexity.}
    \label{fig:sg_dll}
    \end{minipage}
\end{figure*}

In this section, we investigate the relationship between a model's syntactic generalization (SG) ability and its \predictivePower. The SG score is a models' average accuracy across 34 targeted syntactic tests, whose designs are inspired by classic psycholinguistic assessments of human syntactic abilities. Figure \ref{fig:sg_ppl} reproduces Figure 2 from \citeA{hu2020assessing}, which shows the range of SG scores achieved by our models, plotted against each model's test perplexity. \citeA{hu2020assessing} argue that among the range of architectures and training dataset sizes investigated, it is model class, rather than training data size or test perplexity, is the most important determinant of a model's syntactic generalization capabilities. For example, looking at Figure \ref{fig:sg_ppl}, 
the best performing LSTM model (squares) achieves a lower SG score than the lowest performing RNNG models (diamonds). The exception is GPT-2: the GPT-2 model trained on the smallest dataset performs on par with the $n$-gram models; however, the GPT-2 models trained on larger datasets with BPE encoding perform even better than the best performing RNNG models. 

We use SG scores to quantify the degree to which a model has derived human-like syntactic knowledge of language from text. Figure \ref{fig:sg_dll} shows the relationship between models' SG scores and their \predictivePower as $\dll$. We plot this as a residualized regression, testing the relationship between syntactic generalization score and $\dll$ after controlling for the effects of perplexity on both variables.
The x-axis depicts each model's syntactic generalization score residualized with respect to its perplexity (computed within each training vocabulary), and the y-axis shows each model's $\dll$ residualized with respect to its perplexity (again computed within each training vocabulary). The plot thus demonstrates the relationship between the two variables unexplained by the relationship between perplexity and $\dll$.

Many models in this figure show a large amount of variance in residual $\dll$ unexplained by SG score, even when trained on the same dataset. For example, the range of scores achieved by the RNNG BLLIP-XS model overlap with 16/25, or about 64\%, of the other models. We confirm this result quantitatively: in a stepwise regression analysis, SG scores do not significantly improve prediction of $\dll$ over and above perplexity measures of models ($p > 0.26$ for all three corpora).

\section{Discussion}

This paper tested the relationship between language model surprisal estimates and human reading behavior across a broad class of state-of-the-art language models, trained on varying amounts of language data. We confirmed the generally linear relationship between word-level surprisal and human reading time in each of our replications, and discovered that within model architecture, the relationship between a language model's next-word prediction performance and its \predictivePower is mostly monotonic. However, the influence of language model architecture was substantial. Furthermore, the influence of model architecture on \predictivePower is not the same as the influence of model architecture on performance on controlled grammatical tests: we found no clear relationship between the two types of evaluation metrics, once perplexity is controlled (Figure~\ref{fig:sg_dll}).



Our results complement and add to those of \citeA{aurnhammer2019comparing} and \citeA{merkx2020comparing}, who use similar methodology to assess the \predictivePower of Transformers and gated vs.\ simple RNNs. The relatively strong performance of our $n$-gram model accords with \citeauthor{aurnhammer2019comparing}'s \citeyear{aurnhammer2019comparing} finding that simple RNNs, which are more sensitive to local relationships, perform as well as LSTMs and other gated models. Together these results demand a more thorough investigation into the relationship between locality and predictive power. One point of contrast is that \citeA{merkx2020comparing} find no advantage for Transformer models at predicting human reading times in eye-tracking data, although they do find an advantage for self-paced reading. The difference may be due to the assessment metric, testing dataset size, byte-pair encoding or model size (theirs has 2 layers, ours 12). Further investigation is required.

Interpreting our results in light of the findings presented in \citeA{hu2020assessing}, who assess the relationship between perplexity and syntactic generalization abilities, our findings suggest a dissociation between two aspects of cognitive modeling using language models. On one hand, syntactic generalization abilities are largely determined by model architecture, with structurally supervised models and deep Transformers outperforming recurrent neural networks and $n$-gram models. On the other hand, model ability to predict human reading times is determined more by model ability to accurately predict the next word across a range of contexts, not just in specialized syntactic testing. 
For these tasks, model architecture seems to play less of an absolute role, although GPT-2 models trained on larger datasets and enhanced with BPE achieve the highest scores on all three testing corpora. 

The findings presented in this paper suggest that different language comprehension contexts---isolated-sentence reading with controlled materials targeting specific grammatical contrasts, versus reading of more naturalistic materials---bring to the fore different types of human linguistic expectations that are in many cases best captured by different contemporary NLP models.  As new model architectures and training procedures continue to emerge, continued examination of the relationship with psychometric data can help guide the way towards increasingly human-like high-performance computational models of language.



\section{Acknowledgments}

The authors would like to thank the anonymous reviewers for their feedback. J.G.~is supported by an Open Philanthropy AI Fellowship. J.H.~is supported by the NIH under award number T32NS105587 and an NSF Graduate Research Fellowship. R.P.L.~gratefully acknowledges support from the MIT-IBM Watson AI Lab, a Google Faculty Research Award, and a Newton Brain Science Award.

\bibliographystyle{apacite}

\setlength{\bibleftmargin}{.125in}
\setlength{\bibindent}{-\bibleftmargin}

\bibliography{cogsci2020}

\end{document}